\pdfoutput=1
\documentclass[11pt]{article}

\usepackage[preprint]{acl}
\usepackage{float}
\usepackage{multirow}
\usepackage{algorithm}
\usepackage{algorithmic}
\usepackage{times}
\usepackage{latexsym}
\usepackage{amsmath}
\usepackage{booktabs}
\usepackage{subcaption}
\usepackage[T1]{fontenc}
\usepackage[utf8]{inputenc}

\usepackage{microtype}

\usepackage{inconsolata}
\usepackage{graphicx}
\usepackage[normalem]{ulem}
\usepackage{amssymb}

\title{Surgical Feature-Space Decomposition of LLMs: Why, When and How?}

\author{Arnav Chavan$^*$ \\
  Nyun AI, India\\
  \texttt{arnav.chavan@nyunai.com} \\\And
  Nahush Lele$^*$ \\
  Nyun AI, India\\ \And
  Deepak Gupta \\
  Transmute AI Hub, TEXMiN,\\
  IIT (ISM) Dhanbad \\}

\begin{document}
\maketitle
\def\thefootnote{*}\footnotetext{These authors contributed equally to this work. Work done while Nahush was an intern at Nyun AI}
\begin{abstract}
Low-rank approximations, of the weight and feature space can enhance the performance of deep learning models, whether in terms of improving generalization or reducing the latency of inference. However, there is no clear consensus yet on \emph{how}, \emph{when} and \emph{why} these approximations are helpful for large language models (LLMs). In this work, we empirically study the efficacy of weight and feature space decomposition in transformer-based LLMs. We demonstrate that surgical decomposition not only provides critical insights into the trade-off between compression and language modelling performance, but also sometimes enhances commonsense reasoning performance of LLMs. Our empirical analysis identifies specific network segments that intrinsically exhibit a low-rank structure. Furthermore, we extend our investigation to the implications of low-rank approximations on model bias. Overall, our findings offer a novel perspective on optimizing LLMs, presenting the low-rank approximation not only as a tool for performance enhancements, but also as a means to potentially rectify biases within these models. Our code is available at \href{https://github.com/nyunAI/SFSD-LLM}{GitHub}.
\end{abstract}

\section{Introduction}

The mass adoption of Large Language Models (LLMs) in various domains has brought to the forefront the critical need to understand and optimize their structural and functional properties. This understanding is not only crucial from a theoretical point of view, but also essential for practical, real-world applications. LLMs, characterized by their vast size and complexity, present significant challenges in terms of resource utilization and operational efficiency. In this context, model compression has become a crucial topic of research.

Compression of LLMs is not simply a matter of reducing their size, but also about retaining functionality and performance in a more resource-efficient environment. Traditional compression methods, such as quantization, pruning, distillation, low-rank decomposition; have demonstrated considerable success in making LLMs practical. However, parametric reduction of LLMs is non-trivial owing to the massive compute required to re-train or fine-tune compressed models. Therefore, traditional compression methodologies, such as pruning and distillation, undergo substantial performance degradation in a training-free compression regime. We hypothesize that the unique advantage of low-rank decomposition in providing direct control over the low-rank factors contributes to a more effective and training-free compression tendency in LLMs and explore this hypothesis in depth.

In this work, we delve into the low-rank decomposition properties of LLMs. We meticulously examine and surgically decompose the individual layers of these models with the aim of achieving training-free compression while maintaining or, in some cases, enhancing performance. This layer-wise decomposition approach allows for granular analysis of the model's components, leading to a more tailored and effective compression strategy. Furthermore, we extend our study to investigate the effects of low-rank decomposition on the intrinsic biases inherent in LLMs. Bias in language models is a topic of growing concern given the widespread application of these models in society. By examining how decomposition influences these biases, we contribute to the broader conversation on ethical AI and responsible model deployment.

In summary, this work presents a comprehensive study of low-rank decomposition in LLMs, exploring its potential to achieve considerable compression levels and its impact on model performance and biases. Through our empirical analyses, we aim to provide insights that can guide future developments in the field of language model optimization. The contributions of this paper can be summarized as follows:
\begin{itemize}
    \item We introduce Surgical Feature-Space Decomposition (SFSD), a simple method for efficient LLM compression through precise layerwise decomposition of latent features.
    \item Our research empirically demonstrates the effectiveness of reduced-rank approximations on latent features in LLMs, showing its advantages over traditional weight space approximations.
    \item We explore the impact of SFSD on the reasoning performance and perplexity of LLMs, providing a comparative analysis with a recent parametric and training-free structured pruning method.
    \item We additionally study the influence of SFSD on the bias of pre-trained LLMs, quantitative change in sample predictions and general low-rank structures; factors vital for understanding and practical application of SFSD.
\end{itemize}

\section{Related Work}

This work focuses on surgically creating low-rank approximations of the feature space for achieving efficient and effective compression of LLMs. Further, we also study the behavior of the resultant models from a fairness perspective. In this regard, we present below an overview of the existing works related to these areas of research.
% This work tries to provide insights into areas spanning model compression, reduced order modeling, generalization and fairness of language models among others. In this section we provide a concise review of recent progress in these areas of research.

\vspace{0.5em}
\noindent\textbf{Surgical fine-tuning. }Recent works have shown that fine-tuning only specific layers of a deep neural network improves adaptations to distribution shifts \cite{lee2022surgical}. \cite{lodha2023surgical} provided strong empirical results on surgical fine-tuning of the BERT \cite{devlin2018bert} model. Their results show that fine-tuning only specific layers often surpass downstream task performance on GLUE and SuperGLUE \cite{wang2018glue, wang2019superglue} datasets. \cite{shen2021partial} adopt an evolutionary search mechanism to identify layer-specific learning rates and freeze some layers in the process for better performance on few-shot learning tasks. \cite{Park_2024_WACV} perform layer wise auto-weighting using the Fischer Importance Matrix for test-time adaptation. In this process, their method even makes certain layers nearly frozen to mitigate outliers. These works have shown that for any given downstream task, fine-tuning and/or freezing only specific layers lead to better performance as compared to the full model. We argue that not layers, but sub-components in each layer identified through traditional rank decomposition can be surgically eliminated form deep neural networks with minimal loss in performance. 

\vspace{0.5em}
\noindent\textbf{Model compression. }Model compression techniques such as network pruning \cite{frankle2018lottery, liu2018rethinking, tiwari2020chipnet, chavan2022vision}, knowledge distillation \cite{jiang-etal-2023-lion, dasgupta-etal-2023-cost, hsieh2023distilling}, quantization \cite{dettmers2022gpt3, xiao2023smoothquant, lin2023awq}, etc., have been proven instrumental for the practical use of large-scale deep learning models. With the advent rise of LLMs, traditional compression methodologies often fail due to multiple factors including infeasibility for training-aware compression and huge hardware costs among others. Specifically for LLMs, fine-tuning/training-free methodologies have been proposed in recent literature \cite{ma2023llm, dettmers2022gpt3, chavan2024faster}, however they either come with a substantial drop in performance or only work on specific hardware architectures. Additionally most training-free compression methods for LLMs are not data aware making them suboptimal for accurate task-specific compression. In this work, we propose to use feature space decomposition for training-free compression on downstream tasks. Additionally, consistent with previous studies that indicate that model compression serves as an effective regularizer and often improves performance at lower compression levels \cite{frankle2018lottery, chavan2022dynamic}, this work also demonstrates a comparable trend in the context of our proposed methodology. Overall, our work aims to provide a zero-shot and training-free compression method in contrast to existing training-aware methods on smaller models. We perceive quantization as orthogonal to our study and we do not include any LLM distillation due to the heavy fine-tuning requirements of the same which defeats the motivation of this work.

\vspace{0.5em}
\noindent\textbf{Low Rank Approximation. }Using low-rank decomposition to compress traditional convolutional neural networks (CNN) has been widely studied in the literature \cite{denton2014exploiting, jaderberg2014speeding, 7332968, Yu_2017_CVPR}. More recently, \citep{phan2020stable} attempted to stabilize the low-rank approximation by mitigating the degeneracy in decomposition matrices. \cite{idelbayev2020low} proposed a mixed discrete-continuous optimization framework to estimate the optimal layer-wise ranks for compression. All these methods restrict to weight or CNN kernel decomposition, making them oblivious to the data domain, further necessitating the fine-tuning stage of the decomposed networks. Similarly in the domain of NLP, \cite{Hajimolahoseini2021CompressingPL} proposed progressive SVD-based approximation of the GPT-2 model \cite{radford2019language} with re-training to achieve much better compression rates as compared to distillation based compression methods. However, retraining remains pivotal to regaining performance drop during the data-free weight decomposition stage. More recently, \cite{li2023losparse} approximate weight matrices as low-rank and sparse approximations jointly. However, their method requires heavy fine-tuning on each downstream evaluation task. \cite{NEURIPS2021_f56de5ef} proposed a data-aware low rank approximation method for model compression. They employ a dual SVD on model weights and inputs respectively, and propose a closed form solution to the data-aware low rank approximation problem. Their motivation on having data-aware compression closely aligns with ours; however, they show limited evaluation on discriminative language models.

\vspace{0.5em}
\noindent\textbf{Bias and Fairness. }\cite{gallegos2023bias} provide a detailed survey on bias and fairness in large language models. In the context of this work, \cite{ramesh2023comparative} provided an initial study on the impact of model compression on the fairness and bias of language models. However, their study was only restricted to discriminative models and tasks. Similarly, \cite{gonccalves2023understanding} provided solid observations like longer pre-training and larger models have more social biases as compared to the compressed counterparts, among others. \cite{shao-etal-2023-erasure, shao2023gold, Kleindessner2023} have attempted to mitigate model bias by eliminating low-rank components of data and model weights that contribute to a higher bias. In this work, we also focus on understanding the effect of a low-rank approximation on the biases of LLMs.

\section{Low-Rank Decomposition of LLMs}
Before describing low-rank decomposition of LLMs, we first present here a brief understanding of the LLM structure. Any LLM typically involves a deep neural network architecture based on the Transformer model. Without loss of generality, any transformer model has Multi-Head Self Attention (MHSA) and Multi-Layer Perceptron (MLP) blocks repeated across the model depth. We refer to a combination of MHSA and MLP blocks as a single module. Each MHSA block consists of three linear transformation layers corresponding to query, key and value. We denote the weight matrices corresponding to them as $W_q$, $W_k$ and $W_v$ respectively. Additionally, after the attention operation, an output linear transformation layer is present with weight denoted as $W_o$. Similarly, MLP block typically consists of a gate projection layer, an up projection layer and a down projection layer. We denote the weight matrices corresponding to them as $W_g$, $W_u$ and $W_d$ respectively. 

Combining the weights from the MHSA and MLP blocks, the weight space of a LLM with $L$ such repeated blocks can be stated as
\begin{equation*}
\textbf{W} = \bigcup_{l=1}^{L} \textbf{W}_l = \textbf{W}_1 \cup \textbf{W}_2 \cup \ldots \cup \textbf{W}_L 
\end{equation*}
Here, $\textbf{W}_l = \{W_{lq}, W_{lk}, W_{lv}, W_{lo}, W_{lg}, W_{lu}, W_{ld}\}$ denotes the weight space for the $l^{\text{th}}$ block.

\subsection{Weight Space Decomposition}

It refers to taking a simplified approximation of the weight matrices that can lead to reduced computational complexity of the associated mathematical operations. For the reduction of the weight space, we employ Singular Value Decomposition (SVD) on the individual weight matrices. For any particular $W_i \in \textbf{W}$ where $W_i \in \mathbb{R}^{d_2 \times d_1}$, the SVD formulation can be stated as:
\begin{equation}
W_{i} = U \Sigma V^T.
\end{equation}
where $U \in \mathbb{R}^{d_2 \times d_2}, \Sigma \in \mathbb{R}^{d_2 \times d_1}, \text{ and } V^T \in \mathbb{R}^{d_1 \times d_1}$ are the resultant matrices. 

For any desired parametric budget $\beta \in (0,1)$, the total number of final parameters should be $\beta \times d_2 \times d_1$ where $d_2 \times d_1$ are the original number of parameters. This information is then used to calculate the rank of the system that would satisfy the prescribed budget.

For a given rank $r$, we define $U_r \in \mathbb{R}^{d_2 \times r}, \Sigma_r \in \mathbb{R}^{r \times r}, \text{ and } V^T_r \in \mathbb{R}^{r \times d_1}$ as the resultant matrices that approximate $W_i$ as:
\begin{equation}
\tilde{W_{i}} = U_r \Sigma_r V^T_r = W_d W_u
\end{equation}
 where $W_d = U_r \Sigma_r \in \mathbb{R}^{d_2 \times r}$; $W_u = V^T_r \in \mathbb{R}^{r \times d_1}$. Based on this, the relation between the compression budget and system rank can be stated as follows.
\begin{align}
d_2 \times r + r \times d_1 = \beta \times d_2 \times d_1 \implies \nonumber \\
r = \beta \times(d_2 \times d_1) / (d_2 + d_1) = \beta \times \kappa. \label{eq_kappa}
\end{align}
Eq. \ref{eq_kappa} can be used to estimate the rank $r$ for any desired compression budget $\beta$.

\subsection{Feature Space Decomposition}
It is important to note that the above decomposition is not data-aware, it can provide a general approximation of any given set of weights. The distribution of input features of each layer follows a specific pattern, which can be directly influenced by the input samples \cite{schwarzenberg2021studying}. Thus, feature space decomposition can help mitigate errors introduced by general low-rank approximations. Similar to PCA \cite{shlens2014tutorial}, we employ Eigenvalue Decomposition on the covariance matrix of the output features. Assuming $X_i \in \mathbb{R}^{d_1\times D}$ is the input to any $W_i \in \textbf{W}$, where $D$ is the number of calibration data samples; the decomposition can be stated as:
\begin{align*}
Y_{i} = W_iX_i \text{ and } \Sigma_Y = Y_iY_i^T \\
\Sigma_Y = V \Lambda V^T
\end{align*}
where $V \in \mathbb{R}^{d_2 \times d_2}$ and $\Lambda \in \mathbb{R}^{d_2 \times d_2}$ is a diagonal matrix. Assuming that $r$ satisfies the budget constraint $\beta$, $Y_i$ can be approximated as:
\begin{align*}
\tilde{Y_i} = V_r V_r^T W_i X_i = W_d W_u X_i
\end{align*}
where $W_d = V_r \in \mathbb{R}^{d_2 \times r}$ and $W_u = V^T_r W_i \in \mathbb{R}^{r \times d_1}$. To minimize the error introduced by the low-rank approximation, we need to identify highly stable eigenvectors that have a minimum output variance across samples. Highly stable eigenvectors can be replaced by a static bias term, while unstable eigenvectors must be retained. Since eigenvectors are orthonormal; the output can be represented as $Y_i = \sum_{k} V_k^tV_kW_iX_i.$ The output variance of any given eigenvector $V_k$ can be represented as: 
\begin{equation*}
S_k = Var(Y_i|V_k)
\end{equation*}
\begin{equation*}
= \frac{1}{N-1}\sum_{j=1}^{D}(V_k^tV_kW_iX_j - \frac{1}{N}\sum_{j=1}^{D}V_k^tV_kW_iX_j) \\
\end{equation*}

The lower the value of $S_k$, the higher the stability of the $k$-th eigenvector. Co-incidently, $S_k$ is numerically equal to the eigenvalue corresponding to the $k$-th eigenvector. Hence, eigenvectors with low eigenvalues can be replaced by a static bias term, while eigenvectors corresponding to higher eigenvalues are retained.

\vspace{0.5em}
\noindent\textbf{Low-rank bias compensation. }In the case of feature space decomposition, we compensate for the lost information by an additional bias term. Assuming that $V_{r^c}$ denotes the eliminated eigenvectors, we can approximate $Y_i$ by using the orthogornal property of $V$:
\begin{align*}
Y_i = \tilde{Y_i} + V_{r^c} V_{r^c}^T W_i X_i
\end{align*}
We approximate $V_{r^c} V_{r^c}^T W_i X_i$ by taking a mean over the calibration data samples:
\begin{equation}
b_i = V_{r^c} V_{r^c}^T W_i \bar{X_i}
\end{equation}
\begin{equation}
\text{Finally, }\tilde{Y_i} = W_d W_u X_i + b_i
\label{feat}
\end{equation}
This bias compensation helps in approximating the highly stable eigenvectors through a static bias calculated over the calibration data samples without any need to change the original layer. Note that a similar bias compensation cannot be formulated for weight space decomposition as it is data-free, and no mean approximation can be carried out of the stable singular vectors.

\subsection{Surgical Rank Search}
\begin{algorithm}
\caption{SFSD Rank Search}
\label{alg:sfsd}
\begin{algorithmic}[1]
\REQUIRE Pre-trained model $f$, $P$ (Performance Metric), $\beta \in \{0.1, 0.2, \ldots, 0.9\}$ 
\ENSURE $R$ 

\STATE Initialize $R$ as an empty list to store ranks
\FOR{each layer $L \in f$, starting from the last layer}
    \FOR{each budget $\beta_i$ in the set of $\beta$}
        \STATE $r_i$ $\leftarrow$ Determine rank corresponding to $\beta_i$ using Eq. \ref{eq_kappa}
        \STATE $L_i$ $\leftarrow$ Decompose $L$ with rank $r_i$ using Eq. \ref{feat}
        \STATE Overwrite $f$ with $L_i$
        \IF{$f$ meets or exceeds $P$}
            \STATE Append $r$ to $R$
            \STATE \textbf{break} and proceed to the next layer
        \ENDIF
    \ENDFOR
\ENDFOR
\RETURN $R$ 
\end{algorithmic}
\end{algorithm}

\begin{table*}[!ht]
\centering
\caption{Performance of LLaMA-7B model, with weight space and activation space decomposition. Average denotes average performance across the given downstream tasks.}
\resizebox{\textwidth}{!}{%
\begin{tabular}{lcccccccccc}
\toprule
Decomposition & \#Params (B) &\#MACS & BoolQ & PIQA & HellaSwag & WinoGrande & ARC-e & ARC-c & Average \\
\hline
Baseline & 6.7 & 423.93 & \textbf{75.04} & \textbf{78.67} & \textbf{76.22} & \textbf{70.00} & \textbf{72.85} & \textbf{44.88} & \textbf{69.61}  \\ \hline
Feature Space & 5.4 & 339.99 & 74.34 & 74.86 & 66.72 & 67.40 & 66.33 & 39.42 & 64.68\\
Weight Space & 5.4 & 339.99 & 62.20 & 62.57 & 43.91 & 58.80 & 44.95 & 30.03 & 50.41 \\
LLM-Pruner & 5.4 & 339.60 & 57.06 & 75.68 & 66.80 & 59.83 & 60.94 & 36.52 & 59.47 \\
\hline
Feature Space & 3.4 & 215.61 & 62.02 & 61.37 & 34.64 & 56.43 & 40.32 & 28.75 & 47.25 \\
Weight Space & 3.4 & 215.61 & 62.08 & 53.59 & 27.88 & 48.46 & 27.15 & 27.05 & 41.10 \\
LLM-Pruner & 3.4 & 206.59 & 52.32 & 59.63 & 35.64 & 53.20 & 33.50 & 27.22 & 43.58 \\
\bottomrule
\end{tabular}%
}
\label{weightspace}
\end{table*}
We estimate the decomposition ranks for each layer by employing a simple linear search mechanism. For each layer, we search for the lowest possible value of $\beta \in \{0.1,0.2,...,0.9\}$ and the corresponding rank $r$ which satisfies a pre-defined performance metric; more information on the relationship between different values of $\beta$, corresponding rank $r$ and the final budget is provided in the Appendix \ref{sec:rank}. We surgically follow this process for each layer starting from the last layer and sequentially moving to the previous layers. This order is followed throughout our experiments on the basis that the last layers are more susceptible to compression as compared to the earlier layers \cite{ma2023llm}. We employ two types of performance metrics for search, one targeted at preserving the general Wikitext-2 \cite{merity2016pointer} perplexity and another targeted at maximizing the downstream commonsense reasoning task performance. In both cases, we used 20\% data for search and finally evaluate the compressed models on the unseen 80\% data. The exact algorithm is outlined above. Once we surgically decompose the network, each linear layer is replaced by two low-rank linear layers operating sequentially. Thus reducing the parameters and FLOPs of the pre-trained network directly. 
\section{Experimental Analysis}
% In this section, we provide a variety of insights on the decomposition of LLMs, specifically restricting our experiments on LLaMA-7B \cite{touvron2023llama} and Mistral-7B \cite{jiang2023mistral}. We show all our experiments on downstream common-sense reasoning tasks. Exact task details can be found in the Appendix \ref{sec:dataset}. We use a calibration data consisting of 512 samples with a maximum sequence length of 128 across our experiments. This data is randomly sampled from the train splits of the downstream datasets, ensuring zero leakage. It is pivotal to note that results shown on $\sim$7B scale translate much better on large-scale models \cite{frantar2023massive}.
Here, we offer various insights into the analysis of LLMs, focusing specifically on experiments conducted with LLaMA-7B \cite{touvron2023llama} and Mistral-7B models \cite{jiang2023mistral}. Our experiments are centered on downstream common-sense reasoning tasks, with detailed task descriptions available in the Appendix \ref{sec:dataset}. Throughout our experiments, we utilize a calibration dataset comprising 512 samples, each with a maximum sequence length of 128. Importantly, this data is randomly selected from the train splits of the downstream datasets, ensuring no data leakage. It is worth noting that the results demonstrated at the $\sim$7B scale tend to extrapolate more effectively to larger-scale models \cite{frantar2023massive}. Decomposition is carried out on a CPU machine and rank search is done on a single L4 GPU with 24GB of VRAM for faster evaluations during search. It is a possibility that a larger size of calibration data with longer sequence lengths can further provide better low-rank approximations.

\subsection{Weight Space vs. Feature Space}
% First, we present a comparison of performance for weight space and feature space decomposition and establish the choice of decomposition methods for the downstream tasks. For a fair comparison, no surgical strategy is chosen with $\beta$ set to 0.6, and all the linear layers are decomposed to 50\% of the original, starting from the end of the network. This translates into the overall final budget of 80\%. 

First, we compare weight space and feature space decomposition methods and determine the preferred choice for downstream tasks. To ensure fairness in the comparison, no surgical strategy is employed. We employ two parametric budgets of 80\% and 50\% in line with existing works. To achieve a parametric budget of 80\% overall, we decompose the last 12 out of 32 modules with a constant $\beta=0.46$; similarly for 50\% budget, we decompose last 24 of the 32 modules with a constant $\beta=0.33$. Note that the above choices are made to avoid instability in the network due to compression of the layers at the start of the network, as also observed in pruning \cite{ma2023llm}. 

Table \ref{weightspace} presents the performance results for weight space and feature space decomposition. It is clearly evident that feature space decomposition can better retain the performance across downstream tasks. We additionally compare the decomposition methods with LLM-Pruner \cite{ma2023llm} to understand how they fair against structured pruning. From the results shown in Table \ref{weightspace}, it is clear that feature space decomposition is consistently superior over model pruning for the same level of training-free compression. Note that for these cases, $\beta$ is constant across layers, yet feature space decomposition is still able to outperform network pruning by large margins.

\subsection{Task-Specific Decomposition}
\begin{figure*}[!ht]
\centering

% Row 1
\begin{subfigure}{.49\textwidth}
  \centering
  \includegraphics[width=\linewidth]{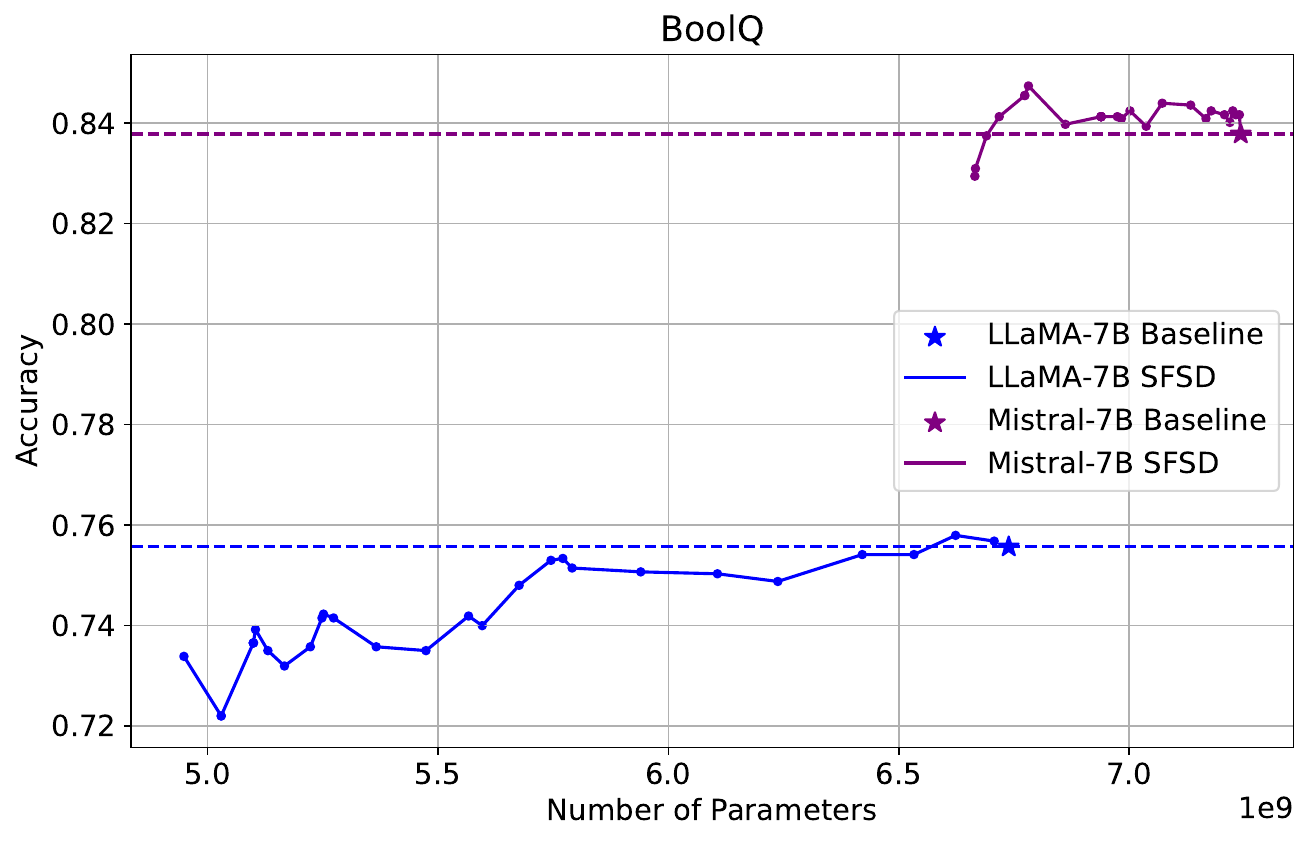}
  % \caption{Caption for image 1}
\end{subfigure}%
%\hfill
\begin{subfigure}{.49\textwidth}
  \centering
  \includegraphics[width=\linewidth]{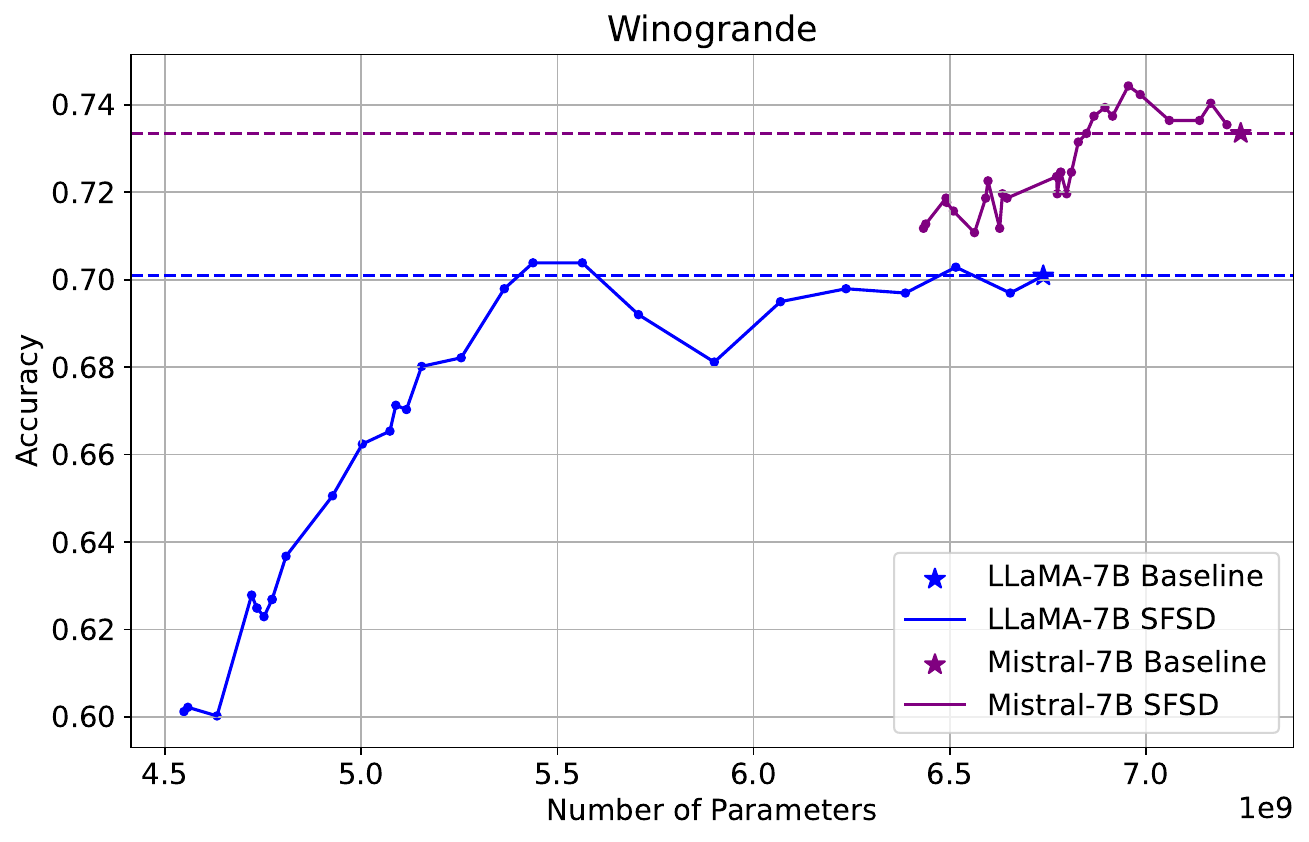}
  % \caption{Caption for image 2}
\end{subfigure}%

\begin{subfigure}{.49\textwidth}
  \centering
  \includegraphics[width=\linewidth]{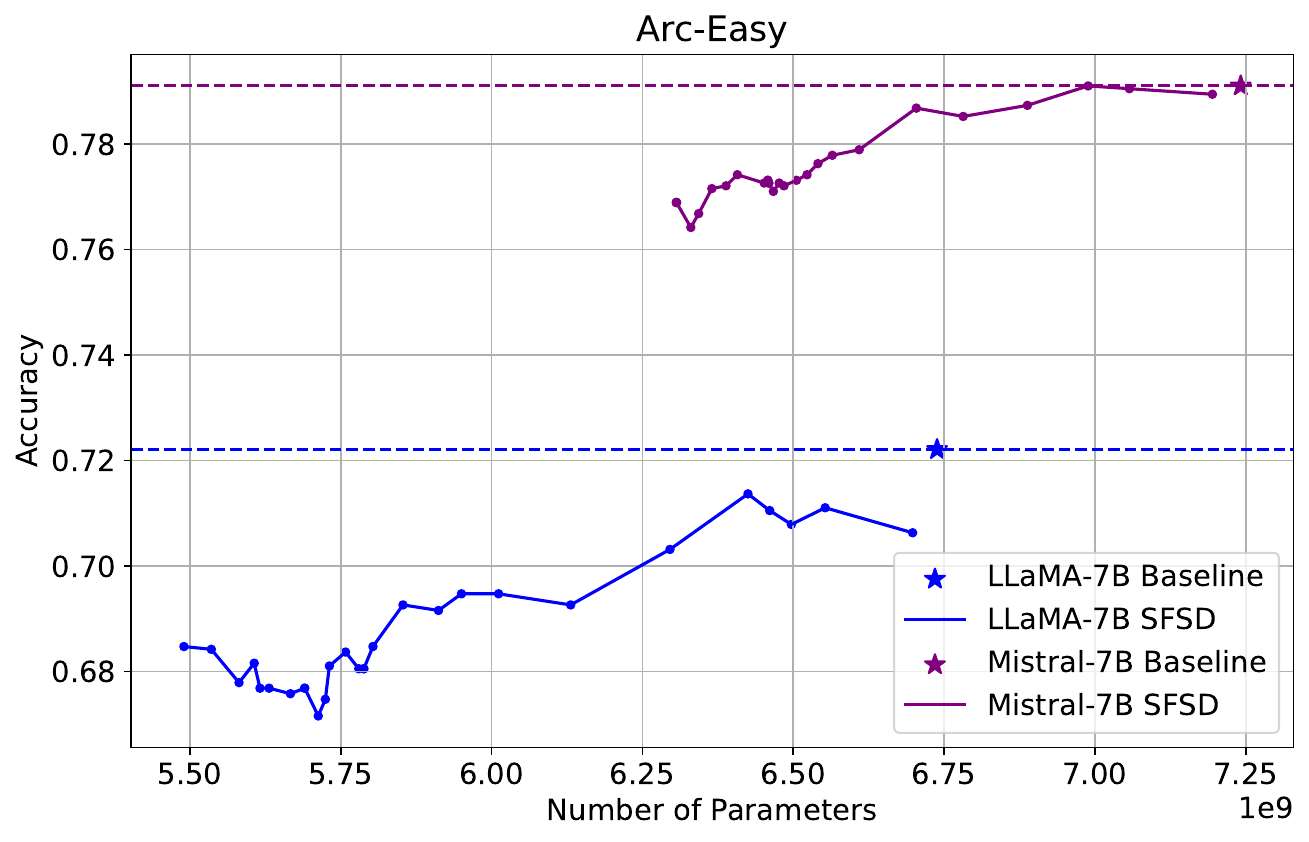}
\end{subfigure}
%\hfill
\begin{subfigure}{.49\textwidth}
  \centering
  \includegraphics[width=\linewidth]{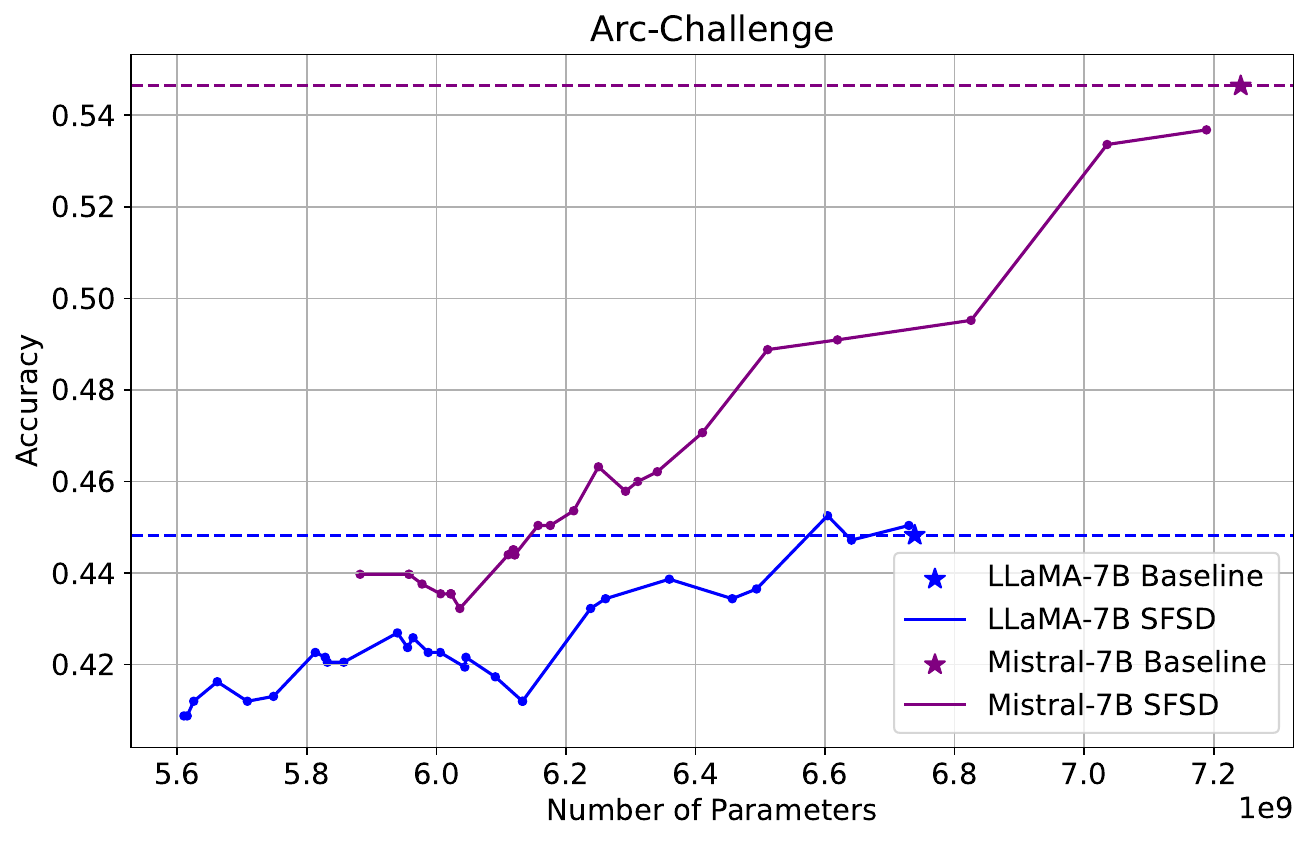}
\end{subfigure}%

\begin{subfigure}{.49\textwidth}
  \centering
  \includegraphics[width=\linewidth]{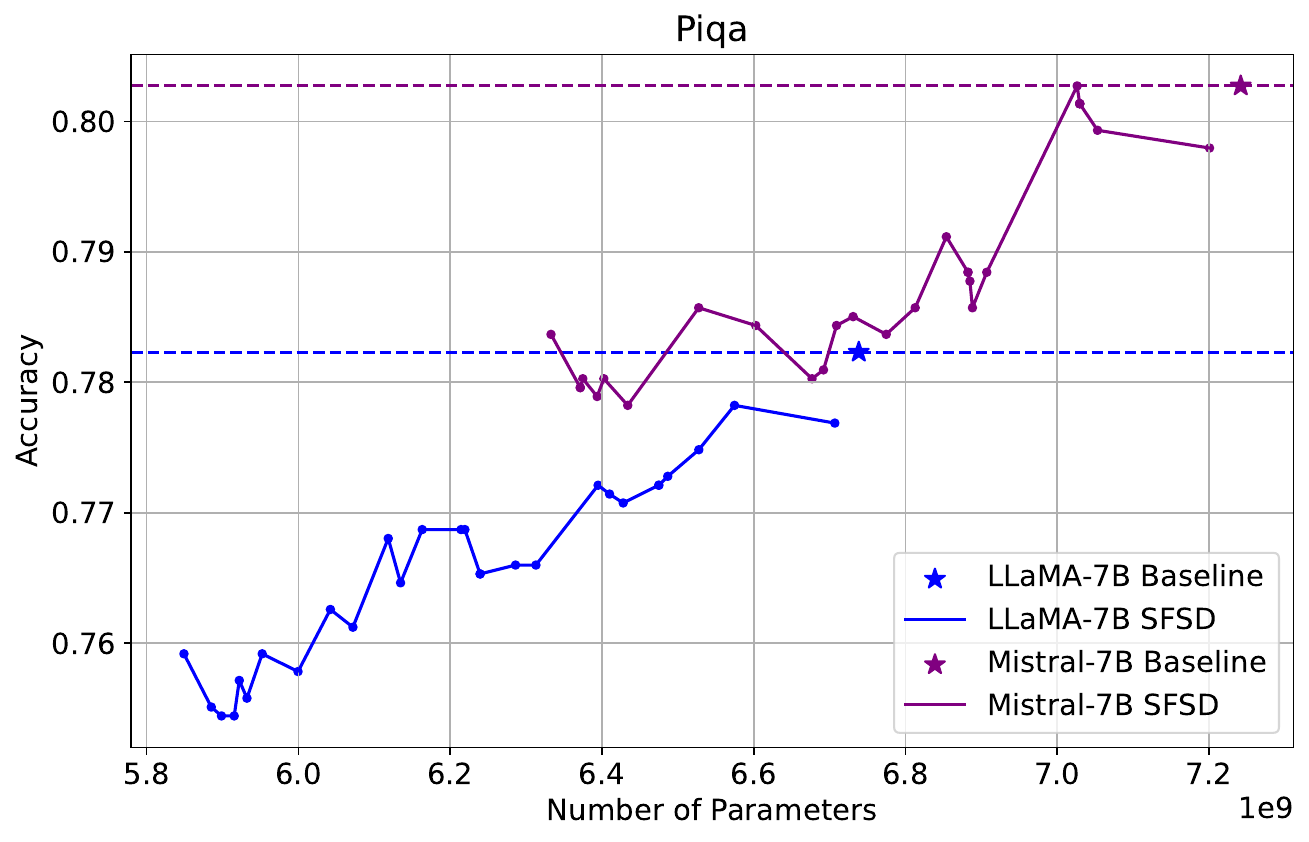}
\end{subfigure}%
%\hfill
\begin{subfigure}{.49\textwidth}
  \centering
  \includegraphics[width=\linewidth]{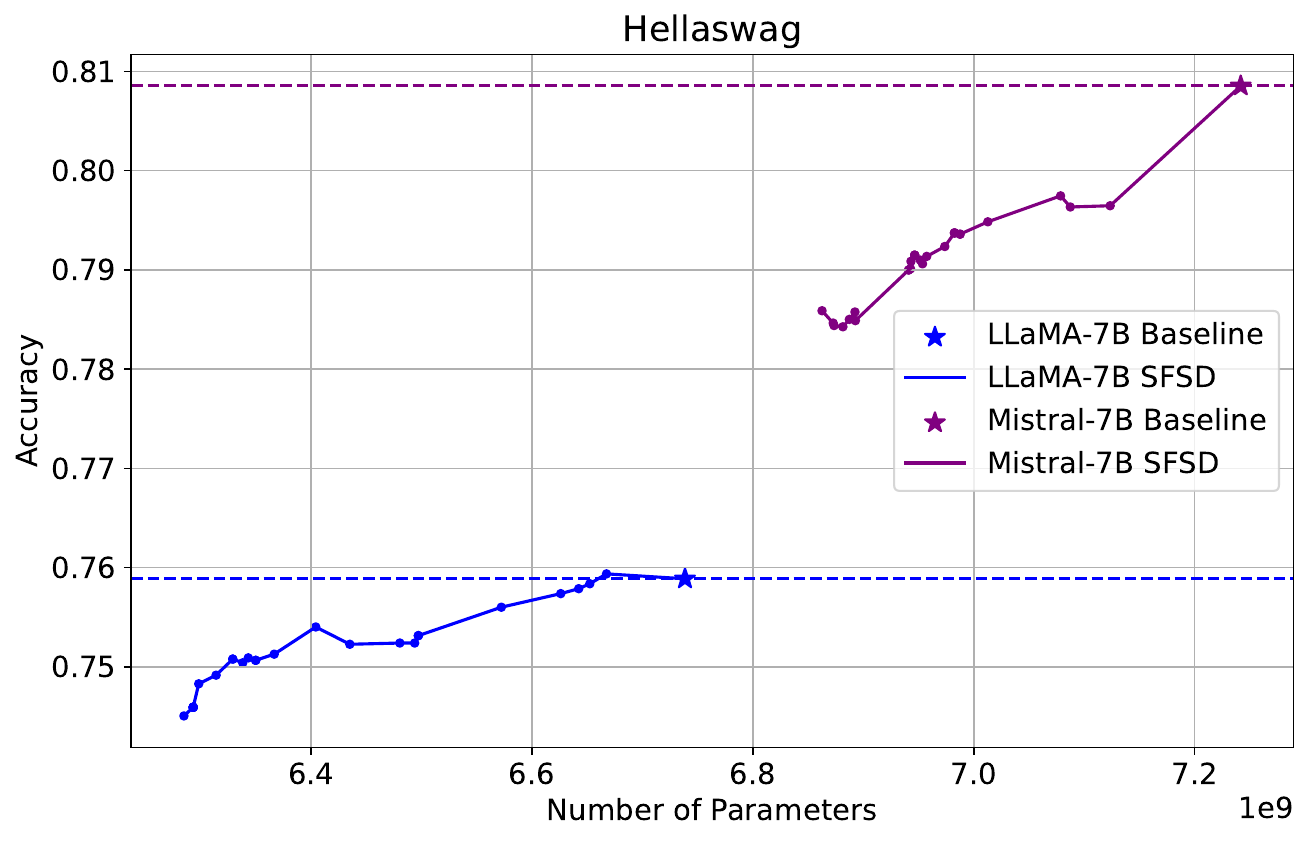}
\end{subfigure}
\caption{Surgical Feature Space Decomposition (SFSD) of LLaMA-7B and Mistral-7B models with task-spcific accuracy used as rank search metric. Horizontal lines indicate the performance of the baseline pre-trained model.}
\label{fig:surgical}
\end{figure*}

Having demonstrated the superiority of feature space decomposition, we employ here an end-to-end surgical decomposition strategy across the complete model. The extensive results are presented in Figure \ref{fig:surgical}. For each downstream task, we search based on the 20\% subset and report performance on the 80\% unseen subset. 

At a relatively lower compression budget, SFSD can retain and even surpass the performance of the base model. This behaviour is consistent across the LLaMA and Mistral models. This makes SFSD a powerful tool for quickly gaining speedups without compromising performance in task-specific environments. Note that owing to the surgical task-specific search mechanism, SFSD automatically estimates and eliminates low-rank eigenvectors per layer depending on the complexity of the target task, while reducing the overall final parameter count dynamically.

Further, we note that the plots shown in Figure \ref{fig:surgical} exhibit a generalization vs. compression trend similar to those of classical pruning literature \cite{frankle2018lottery}. In most cases, models generalize better than the baseline model at minimal compression levels and undergo performance degradation at higher levels of compression. Note that the proposed approach operates without training and gradients, with the potential for lost performance to be regained through fine-tuning the preserved eigenvectors. However, fine-tuning the resultant models is outside the scope of this study.

\subsection{Surgical Decomposition using Perplexity}
\begin{figure*}[!ht]
\centering

% Row 1
\begin{subfigure}{.48\textwidth}
  \centering
  \includegraphics[width=\linewidth]{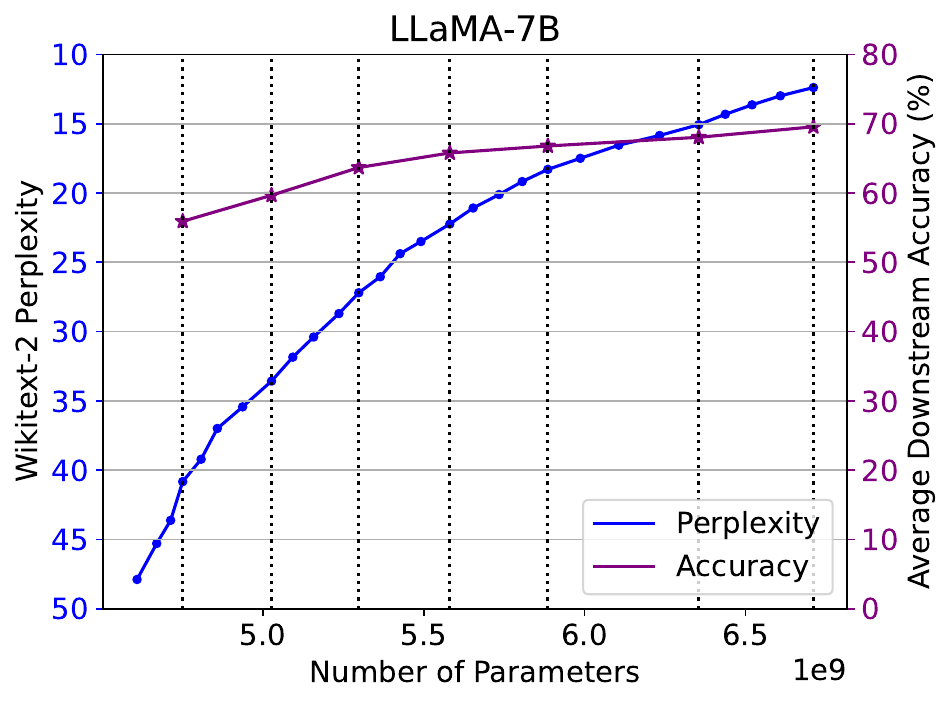}
  % \caption{Caption for image 1}
\end{subfigure}%
%\hfill
\hspace{1em}
\begin{subfigure}{.48\textwidth}
  \centering
  \includegraphics[width=\linewidth]{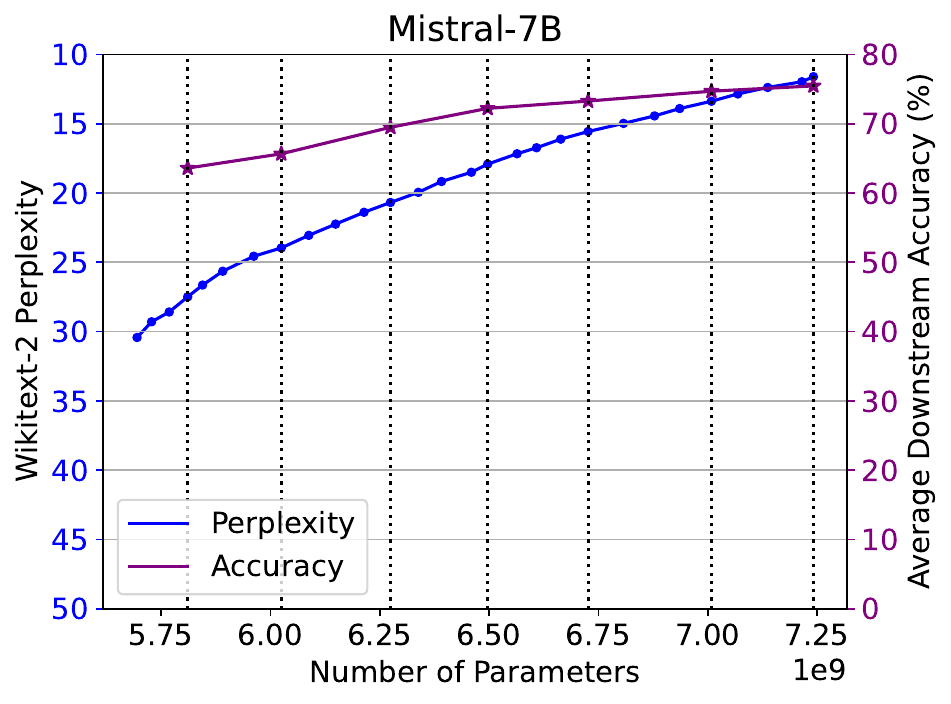}
  % \caption{Caption for image 2}
\end{subfigure}%
\caption{Surgical Feature Space Decomposition (SFSD) of LLaMA-7B and Mistral-7B models with perplexity used as the rank search metric. The decomposed models are evaluated at regular interval on commonsense reasoning tasks and average accuracy is reported.}
\label{fig:ppl}
\end{figure*}

We have demonstrated above the efficacy of SFSD for task-specific compression. Next, we investigate, how the compressed models obtained from SFSD fair when no task-specific information is employed. We present here insights obtained from SFSD-based compression of the models using Wikitext-2 \cite{merity2016pointer} perplexity as the performance metric. Resultant models are then evaluated on commonsense reasoning tasks and an average performance over the six chosen tasks is presented.

The results for LLaMA-7B and Mistral-7B are presented in Figure \ref{fig:ppl}. In line with existing works, the perplexity undergoes substantial degradation as the compression level increases. However, it is noteworthy that average downstream task performance does not degrade substantially. This shows that even if the perplexity deteriorates, the model can retain common sense reasoning to a reasonable extent. Note that across different parameter counts, SFSD beats LLM-Pruner substantially on average downstream task performance; 65.4 \% vs. 59.5 \% at 5.4B parameter count (80\% budget) on the LLaMA-7B model. Mistral-7B exhibits a slower parameter reduction and hence a higher overall performance. The relative drop in perplexity and average accuracy is similar to LLaMA-7B considering the parameter reduction.

\subsection{Effect on Model Bias}
\begin{table}[ht]
\centering
\caption{ICAT Score for baseline and SFSD models. Budget indicates the overall parametric budget post decomposition. A higher ICAT score indicates a less biased model.}
\resizebox{0.48\textwidth}{!}{%
\begin{tabular}{@{}lcccccc@{}}
\toprule
Model          & Budget & Gender & Profession & Race  & Religion & Overall \\ \midrule
               & 100\%  & 56.85  & 66.75      & 60.81 & 75.02    & 63.14   \\
\multirow{-2}{*}{LLaMA-7B} & 90\%   & 58.50  & 68.08      & 62.60 & 78.85    & 64.84   \\
               & 80\%   & 59.31  & 70.50      & 65.73 & 78.30    & 67.28   \\ \midrule
               & 100\%  & 52.03  & 66.43      & 62.63 & 79.79    & 63.45   \\
\multirow{-2}{*}{Mistral-7B} & 90\%   & 54.93  & 68.22      & 62.94 & 87.03    & 64.87   \\
               & 80\%   & 59.05  & 68.12      & 65.99 & 84.68    & 66.90   \\ \bottomrule
\label{bias}
\end{tabular}}
\end{table}

For a holistic evaluation, we also analyse the change in model's bias post SFSD. We evaluate the baseline model and the compressed models on StereoSet \cite{nadeem2020stereoset} - a standard benchmark to measure stereotypical biases in pre-trained language models. We use the generic perplexity-based SFSD models as discussed in the previous section. The Idealized Context Association Test (ICAT) scores are presented in Table \ref{bias}. A higher ICAT score indicates a less biased model. 

It is noteworthy that SFSD substantially reduces bias in the pre-trained LLMs across multiple stereotypes. This observation is consistent across LLaMA-7B and Mistral-7B. This makes SFSD an attractive choice for compression since it reduces intrinsic bias in LLMs in contrast to distillation and pruning which have been shown to increase the bias in pre-trained language models \cite{ramesh2023comparative}.

\section{Discussion and Analysis}
\subsection{Decomposed Model Visualisation}
\begin{figure*}[!ht]
\centering
  \centering
  \includegraphics[width=0.99\linewidth]{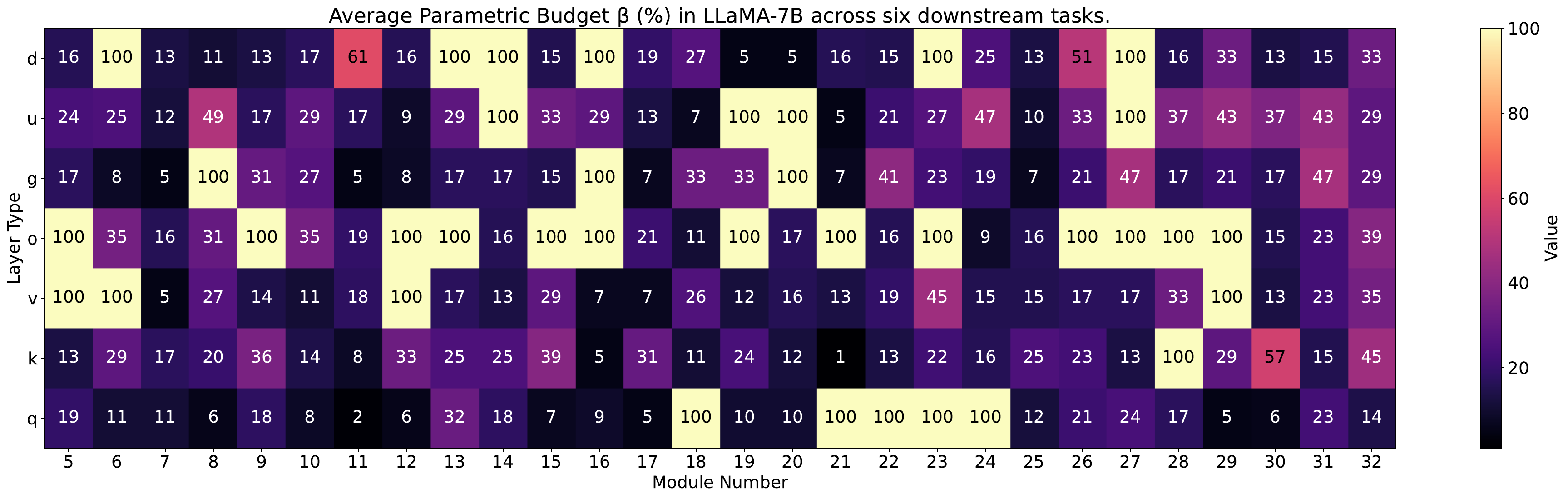}

\caption{Final Parametric Budget $\beta$ averaged across six commonsense reasoning tasks. 100\% indicates an intact layer exactly similar to the pre-trained model. LLaMA-7B consists of a total of 32 modules; with each module having query (q), key (k), value (v), output (o), gate (g), up (u) and down (d) projection layers.}
\label{fig:vis}
\end{figure*}
To further study the low-rank structures in SFSD models, we plot the final parameter budget $\beta$ averaged across the six surgical searches shown in Figure \ref{fig:surgical}. A detailed visualization is provided in Figure \ref{fig:vis}. The module number indicates the location of each module in the network architecture, and the number inside each grid indicates the \% of parameters retained by SFSD as compared to the original pre-trained model parameters. 

Some intrinsic patterns can be seen across datasets, and specifically some layers are consistently intact with respect to the pre-trained model. These layers are pivotal to model performance, hence SFSD prefers to retain these layer completely. We also see some layers undergoing extreme rank reduction, and the average compression across the datasets is relatively very high for them. We observe that compared to MLP layers, attention layers (query, key and value) are more prune to compression through SFSD. More specifically, query projection either undergoes substantial compression or stays completely intact. Additionally, the output projection layer in MHSA undergoes minimum compression and stays intact almost half the times.

\subsection{Learning - Unlearning in SFSD}
\begin{table}%[!ht]
\centering
\caption{Accuracy and disagreement between the 90\% budget model (LLaMA-7B) and the baseline predictions across multiple datasets. Disagreement refers to \% data points whose predictions differ between the pre-trained and the SFSD models.}
\resizebox{0.48\textwidth}{!}{
\begin{tabular}{lcccc}
\toprule
Dataset & SFSD (\%) & Pre-trained (\%) & Delta (\%) & Disagree. (\%) \\
\midrule
PIQA & 77.26 & 78.67 & -1.41 & 5.33 \\
Winogrande & 69.53 & 70.01 & -0.48 & 9.16 \\
BoolQ & 73.24 & 75.05 & -1.81 & 8.41 \\
ARC-e & 72.90 & 75.38 & -2.48 & 8.80 \\
ARC-c & 40.35 & 41.80 & -1.45 & 10.24 \\
Hellaswag & 54.37& 56.96 & -2.56 & 4.61 \\
\bottomrule
\end{tabular}}
\label{disagreement}
\end{table}

To study the drift in learning mechanism post SFSD, we analyse the disagreement between baseline pre-trained model and the corresponding perplexity-based SFSD model. The disagreement and the difference in performance of the two models are presented in Table \ref{disagreement}. It is surprising to note that the disagreement is much higher than the performance difference, indicating that SFSD undergoes substantial learning and unlearning process. Specifically, for harder tasks like Arc-Challenge, the disagreement is notably higher indicating that SFSD is able to recover representations inherently hidden in the network. However, this comes at the cost of unlearning representations already present in the pre-trained model. By designing search metrics targeted at unlearning or debiasing, SFSD unlocks a promising direction to recover hidden representations present in the network and mitigate targeted biases.

\subsection{Computational Efficiency}
On average, it takes \textasciitilde 13 seconds for a single layer of LLaMA-7B to decompose. Decomposing all the 224 layers takes roughly \textasciitilde 45 minutes. These numbers are reported on a 48 Core 128GB CPU-only machine, thus eliminating the need for a GPU altogether for decomposition. In the case of the rank search mechanism, we employ a single 24GB L4 GPU which is 2-3$\times$ slower and runs at a lower batch size than a 40/80GB A100 machine. Nevertheless, rank search depends on the target data size, with the smallest dataset - Winogrande taking \textasciitilde 8 hours while the largest dataset - Hellaswag takes \textasciitilde 36 hours for the complete rank search. Additionally, not all layers require rank search and even a search-free uniform rank across layers performs better than LLM-Pruner.

\subsection{The Why, When and How of SFSD}
\noindent\textbf{Why? }SFSD offers an alternative and efficient method for model compression without the need for fine-tuning. We demonstrate that feature space decomposition is superior over weight space decomposition, and SFSD performs feature decomposition in a very effective manner. SFSD also outperforms current structured pruning methods on various commonsense reasoning tasks. Beyond the standard performance measures, the resultant compressed models obtained from SFSD exhibit relatively lower intrinsic model bias, with regard to ethical concerns. The multifold advantages of SFSD over other compression methods clearly put it forward as a preferred choice for building efficient LLMs.

%The primary motivation for this approach is to provide an alternative and efficient method for model compression without the need for fine-tuning. The technique of feature-space decomposition offers a better approximation compared to weight-space decomposition in a fine-tuning free setup. It's shown to outperform current structured pruning methods on various commonsense reasoning tasks. Additionally, a significant benefit of this approach is the reduction in intrinsic model bias post decomposition, addressing ethical concerns in AI.

\noindent\textbf{When? }SFSD presents numerous advantageous scenarios where its utility becomes very clear. Firstly, its notably lower computational complexity renders it as the preferred choice in contexts where extensive re-training or fine-tuning of models isn't feasible. Despite the potential memory constraints of full-scale decomposition, SFSD operates on a layerwise basis, mitigating memory load significantly. Moreover, SFSD proves beneficial when the primary objective is to reduce model size without sacrificing performance on reasoning tasks. Additionally, SFSD demonstrates marginal performance enhancements even at modest compression levels, making it a viable option for post-processing to boost model performance. Lastly, SFSD serves a crucial role in the development of compressed LLMs, inherently mitigating model bias and addressing ethical considerations. 
% This method is also valuable when addressing the issue of bias in pre-trained models, as it has consistently shown to reduce model bias.

\noindent\textbf{How? }In simple terms, SFSD implemented layerwise eigenvalue decomposition in the feature space, achieving bias compensation and low-rank approximation. The approach involves surgical feature-space decomposition, which enables the extraction of models with varying parameters from a pre-trained LLM while minimizing performance drops. The decomposition is started from the last layer and moved iteratively towards the first. Detailed description has been presented in \mbox{Algorithm \ref{alg:sfsd}}.

% This method employs Eigenvalue Decomposition, akin to PCA, to achieve bias compensation and low-rank approximation. It surgically decomposes the model layers, starting from the last layer and moving sequentially to previous layers.

With our comprehensive exploration of SFSD's merits in terms of efficiency and ethical AI development, we've firmly established its value in constructing effective LLMs. Its adaptability to various contexts, including memory limitations and ethical considerations, underscores its significance and potential impact. Consequently, SFSD emerges as a promising method for compression of LLMs without the need for extensive training, opening avenues for further research in efficient and ethically conscious model optimization.

\section{Limitations}
While we have sufficiently demonstrated the efficacy of SFSD in compressing and building efficient LLMs, there are still a few limitations that need to be overcome before SFSD can be looked at the preferred full-blown solution for improving LLMs from the efficiency perspective. 

First is the scale of experimentation. While we have observed ourselves from very early experiments that SFSD works well on LLMs with parameters above 20B, the extensive experimentation presented in the paper is on 7B models, and for unhindered generalization, additional experiments with more parameters might be needed.

Next, our rank search strategy is currently surgical, and it is not clear yet whether it exploits the best out of the proposed method. This might be clearly limiting the extent of compression that can be achieved with minimal performance drop, and a better choice in terms of a first-principled approach might be able to develop more efficient LLMs.

\bibliography{acl_latex}
% \clearpage
% \newpage
\appendix

\section*{Appendix}
\begin{table*}[!htbp]
  \centering
  \begin{tabular}{ccccccc}
    \toprule
    Dataset & BoolQ & PIQA & HellaSwag & WinoGrande & ARC-e & ARC-c \\
    \midrule
    Search Split (20\%) & 654 & 367& 2008& 253& 475& 234\\
    Eval. Split (80\%) & 2616 & 1470& 8034& 1013 & 1901& 938\\
    \bottomrule
  \end{tabular}
  \caption{Distribution of sample data used for search and evaluation on task-specific SFSD models.}
\end{table*}

\label{sec:appendix}

\section{Dataset Descriptions}

\label{sec:dataset}
In our analysis, we focus on two sets of data: the calibration dataset selection and the test split of the dataset which is used for the surgical rank selection process.
\subsection{Calibration Dataset}

The model's performance is dependent on the choice of the calibration dataset, as it directly affects the computation of the covariance matrix through the activations of samples, which is subsequently utilized in the eigenvalue decomposition process. We experimented with a variety of calibration datasets including the six common-sense reasoning tasks (BoolQ\cite{clark-etal-2019-boolq}, PIQA\cite{bisk2020piqa}, Arc-Challenge\cite{clark2018think}, Arc-Easy\cite{clark2018think}, Winogrande\cite{10.1145/3474381}, Hellaswag\cite{zellers2019hellaswag}), WikiText-2 \cite{merity2016pointer}, and a combination of all the six aforementioned common-sense reasoning tasks wherein each batch in the calibration dataset contains an equal number of samples from each task. The samples for the calibration dataset in each setting are chosen from a distinct set, separate from the split used for model evaluation and the presentation of our results. Thus we ensure that there is no data leak between the calibration dataset and the evaluation split. In our initial experiments we observed that the combination dataset has superior generalization on downstream tasks as compared to a task-specific calibration dataset, hence all our subsequent experiments as well as the results mentioned in this paper use models decomposed on the basis of the combination dataset. Further, the superior generalization of the model with the combination dataset can be attributed to the fact that our approach to obtaining low-rank matrices benefits from greater variability among the samples within the calibration batch.
\subsection{Surgical Rank Search Dataset}
In the task-specific compression for common sense reasoning tasks, the test dataset is divided into two parts: one containing 20\% of the samples and the other comprising 80\% of the samples. In the rank selection process for a specific layer,the objective is to identify the minimum rank that ensures performance maintenance across the chosen 20\% validation split. The reported results for all task specific compression experiments are on the disjoint 80\% split of the test dataset. \\
Additionally, the common sense reasoning scores for a model compressed based on perplexity are derived from the evaluation of the model on the complete test split of the corresponding common sense reasoning task.

\section{Layerwise Rank and Budget}
\label{sec:rank}
Assuming a set of layerwise budgets $\beta \in \{\beta_1, \beta_2, ..., \beta_N\}$ where N is the total number of linear layers. The final parametric budget can be estimated by $\frac{\sum_{i=1}^{N}\beta_ip_i}{\sum_{i=1}^{N}p_i}$, where $p_i$ denotes number of parameters in the $i^{th}$ layer. Equation \ref{eq_kappa} can be directly used to estimate the rank of any particular layer given a value of $\beta$.

\end{document}